\documentclass[runningheads]{llncs}
\usepackage{fix-cm}
\usepackage[T1]{fontenc}
\usepackage{amsmath,amssymb,amsfonts}
\usepackage{algorithmic,caption}
\usepackage{graphicx}
\usepackage{textcomp}
\usepackage{multirow}
\usepackage{booktabs}
\usepackage{stfloats}

\usepackage[section]{placeins}
\usepackage{subcaption} 
\usepackage{graphicx}
\usepackage{caption}
\usepackage{float}
\usepackage{xcolor}
\def\BibTeX{{\rm B\kern-.05em{\sc i\kern-.025em b}\kern-.08em
    T\kern-.1667em\lower.7ex\hbox{E}\kern-.125emX}}
\usepackage{graphicx}

\begin{document}
\title{OptiModNet: A UNet-Transformer Hybrid with Grouped-Query and Channel Attention for Optic Disc and Cup Segmentation}
%
%
\author{Soumili Ghosh\inst{1}\orcidID{} \and
Debapriya Roy\inst{2}\orcidID{0000-0001-8657-3130}\and
Aryan Das\inst{3}\orcidID{0000-0003-0439-7351} \and
Bikash Santra\inst{4}\orcidID{0000-0002-6833-140X}}
\authorrunning{Ghosh et al.}
%
\institute{IEM Saltlake,India
\email{ghoshsoumili03@gmail.com}\\
\and 
Techno Main Salt Lake, India 
\email{debapriyakundu1@gmail.com}\\
\and
VIT Bhopal University, India
\email{aryan.das2021@vitbhopal.ac.in}\\
\and IIT Jodhpur, India
\email{bikash@iitj.ac.in}}
\maketitle              
%

\begin{abstract}
Precise segmentation of the optic disc and cup is critical for the early detection and diagnosis of glaucoma. However, achieving consistently high performance across datasets while maintaining low computational requirements remains a significant challenge. In glaucoma detection, low-computation methods are crucial for enabling rapid, large-scale screening and facilitating deployment in resource-limited clinical environments. While deep learning models such as UNets, Vision Transformers (ViTs), and Diffusion models have demonstrated strong segmentation performance but these methods often come with substantial computational overhead. UNets are efficient at capturing local features but are limited in modeling global contextual information. Conversely, ViTs excel at long-range dependency modeling but are computationally intensive. Hybrid architectures, such as UNetR, which combine transformer-based encoders with UNet-style decoders, have shown improved performance but while incurring additional complexity. Considering these, in this work, we propose OptiModNet, a light weight novel hybrid architecture tailored for optic disc and cup segmentation. The model integrates diverse attention mechanisms at multiple stages of the network to enhance both local and global feature representation. We include an Aggregated Pyramid Loss that supervises predictions at multiple decoder depths, to promote better gradient flow and structural consistency. We evaluate OptiModNet on the REFUGE2 dataset for both optic disc and cup segmentation tasks. Our method achieves state-of-the-art performance, exceeding existing approaches by over 2.5\%, while maintaining high efficiency with only 3.73 GFLOPs and 1.93M parameters. The code is available at \url{https://github.com/SG1947/OptiModNet}.
\end{abstract}

\section{Introduction}
\begin{figure}{}
    \centering
    \includegraphics[width=0.8\linewidth]{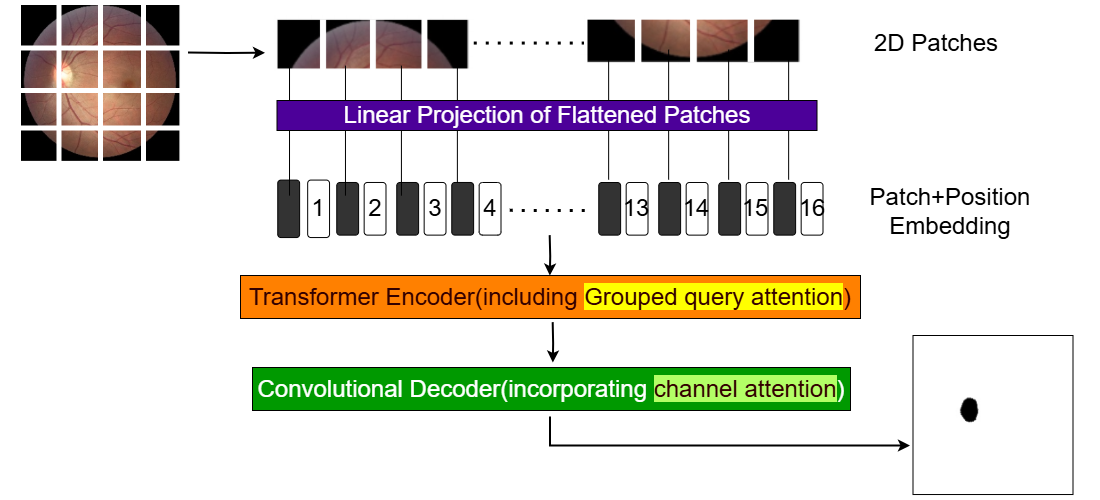}
    \caption{Overview of the proposed OptiModNet.
}
    \label{fig: overview}
\end{figure}

Glaucoma is a major cause of irreversible blindness worldwide and it's early stages are often symptomless. Accurate analysis of retinal fundus images plays a vital role in its detection and monitoring. In particular, segmentation of the optic disc (OD) and optic cup (OC) is a critical step in computing the cup-to-disc ratio (CDR), which is an important biomarker for glaucoma screening and progression assessment. Automated OD/OC segmentation facilitates large-scale screening programs and reduces inter-observer variability in clinical settings. However, OD/OC segmentation is challenging due to factors such as low contrast between cup and disc boundaries, peripapillary atrophy (PPA), blood vessel occlusion at the optic nerve head, and variability in image quality across acquisition devices. Over the years, significant advances have been made in retinal image segmentation, leveraging deep learning architectures to address these challenges. Approaches based on UNet, Vision Transformers (ViTs), and generative models have achieved promising results OD/OC segmentation.

In the last decade, UNet-based models have been widely adopted for OD/OC segmentation due to their strong local feature learning capabilities. Yet, their intrinsic limitation lies in modeling long-range dependencies~\cite{ronneberger2015u}, which are essential for capturing the global anatomical context of the optic nerve head. Conversely, ViTs effectively capture global relationships but require large annotated datasets and incur high computational costs, as computational complexity of the self-attention mechanism increases quadratically with the number of image patches.~\cite{dosovitskiy2020image}. This makes their deployment in high-resolution fundus imaging resource-intensive. Hybrid architectures like TransUNet~\cite{chen2021transunet} and UNetr~\cite{hatamizadeh2022unetr} combine UNet’s local representation learning with transformer-based global context modeling, showing improved performance for OD/OC segmentation. Nevertheless, they inherit the computational inefficiencies of vanilla Transformers. SwinBTS~\cite{jiang2022swinbts} and Swin-UNetr~\cite{hatamizadeh2021swin} address some of these challenges with shifted window mechanisms, yet remain memory-intensive for full-resolution fundus images. Boundary-aware models such as BEAL~\cite{wang2019boundary} and BAT~\cite{wang2021boundary} have enhanced delineation of OD/OC boundaries, but at the cost of additional computational complexity. In addition, diffusion-based approaches like MedSegDiff~\cite{wu2024medsegdiff} and EnsemDiff~\cite{wolleb2022diffusion} have recently demonstrated competitive results for medical image segmentation, but their high training and inference costs make them less suited for real-time clinical workflows.

To overcome these segmentation limitations, we introduce a hybrid UNetR-based architecture that incorporates grouped-query attention in the encoder~\cite{ainslie2023gqa} and channel attention in the decoder~\cite{hu2018squeeze}. While UNetR has shown strong performance in 3D medical segmentation, our 2D adaptation is tailored for fundus images, enabling efficient extraction of both global context and fine boundary details. We also propose a new loss function, Aggregated Pyramid Loss, that enforces supervision at multiple decoder stages, improving gradient flow and enhancing the segmentation of fine OD/OC boundaries. Through extensive experiments on public OD/OC segmentation benchmarks, we demonstrate that our method achieves state-of-the-art accuracy while significantly reducing computational overhead. Our contributions can be summarized as the following:

\begin{itemize}
\item We incorporate grouped query attention into the encoder, reducing computational complexity while effectively modeling long-range dependencies.
\item We incorporate channel attention in the decoder to focus on more informative feature channels, improving boundary delineation between the optic disc and cup.
\item We introduce a novel loss function, Aggregated Pyramid Loss, supervising multiple decoder stages to enhance gradient flow and fine detail segmentation.
\item The proposed model builds upon the complementary strengths of UNet, Transformers, and attention mechanisms to achieve improved OD/OC segmentation accuracy with fewer parameters.
\end{itemize}
\section{Literature Survey}
\label{sec:lit_survey}

Medical image segmentation has advanced significantly with encoder–decoder architectures, notably UNet and its numerous variants. For example, ResUNet~\cite{diakogiannis2020resunet} enhances feature reuse via residual connections, while ARU-GD~\cite{maji2022attention} integrates attention mechanisms to improve localization precision. nnUNet~\cite{isensee2021nnu} offers a self-configuring framework that adapts network architecture, training schemes, and preprocessing steps automatically. BAT~\cite{wang2021boundary} introduces boundary-aware modules to enhance edge precision, and DAGAN~\cite{yang2017dagan} leverages adversarial training to improve robustness and segmentation quality.

Vision Transformers (ViTs) have recently emerged as a promising paradigm for segmentation tasks, capable of capturing global contextual relationships through self-attention, which CNNs—focused primarily on local receptive fields—often miss. The Vision Transformer (ViT)\cite{dosovitskiy2020image} processes images as sequences of patches, proving effective in medical imaging applications. The Swin Transformer\cite{liu2021swin} addresses computational constraints through hierarchical feature extraction with shifted windows. Hybrid architectures such as TransUNet~\cite{chen2021transunet} integrate transformer layers into a UNet encoder, while UNetr~\cite{hatamizadeh2022unetr} and Swin-UNetr~\cite{hatamizadeh2021swin} adopt Swin Transformer designs. Other models like FAT-Net~\cite{wu2022fat} and Swin-UNet~\cite{cao2022swin} focus on fusing local CNN features with global transformer cues. For volumetric segmentation, TransBTS~\cite{wenxuan2021transbts} and SwinBTS~\cite{jiang2022swinbts} combine 3D CNNs with transformer encoders.

Diffusion-based approaches have recently gained traction. EnsemDiff~\cite{wolleb2022diffusion} and SegDiff~\cite{amit2021segdiff} apply score-based models for structured prediction. MedSegDiff~\cite{wu2024medsegdiff} and MedSegDiff-V2~\cite{wu2024medsegdiffv2} leverage denoising diffusion processes to generate sharper, more consistent outputs. These models show strong performance but often increase computational cost. Combining Vision Transformers, UNet, and attention methods~\cite{ainslie2023gqa} creates a powerful tool for medical imaging.

Within the specialized field of OD and OC segmentation, accurate boundary delineation is crucial for computing the CDR, an important biomarker for glaucoma diagnosis. Early works like Sevastopolsky~\cite{sevastopolsky2017optic} adapted UNet for fundus images, while JointRCNN~\cite{jiang2019jointrcnn} introduced region-based convolutional neural networks for simultaneous OD and OC segmentation. Robustness improvements were demonstrated in Yu et al.\cite{yu2019robust} with a deep learning framework capable of handling variable image qualities, and Liu et al.\cite{liu2019spatial} proposed a spatial-aware network for improved combined segmentation. Later, attention-based strategies such as Zhao et al.\cite{zhao2021application} incorporated transfer learning into Attention U-Net for better generalization, while Pachade et al.\cite{pachade2021nenet} leveraged a lightweight Nested EfficientNet with adversarial learning to reduce computational cost without sacrificing accuracy. Transformers have also entered the OC/OD segmentation space, with works like Yi et al.\cite{yi2023c2ftfnet} introducing coarse-to-fine transformer networks for high-precision results and Yan et al.\cite{yan2024mrsnet} combining large-kernel residual convolution with self-attention for consistent segmentation across datasets. Hybrid architectures combining Vision Transformers (ViTs) and UNet have been explored to address the limitations of each while leveraging their complementary strengths~\cite{chen2021transunet,cao2022swin}. Our method falls within this category but emphasizes a lighter design with improved performance across diverse OC/OD datasets.
\section{Method}
\label{sec: method}
\begin{figure*}
    \centering
    \includegraphics[width=1\linewidth]{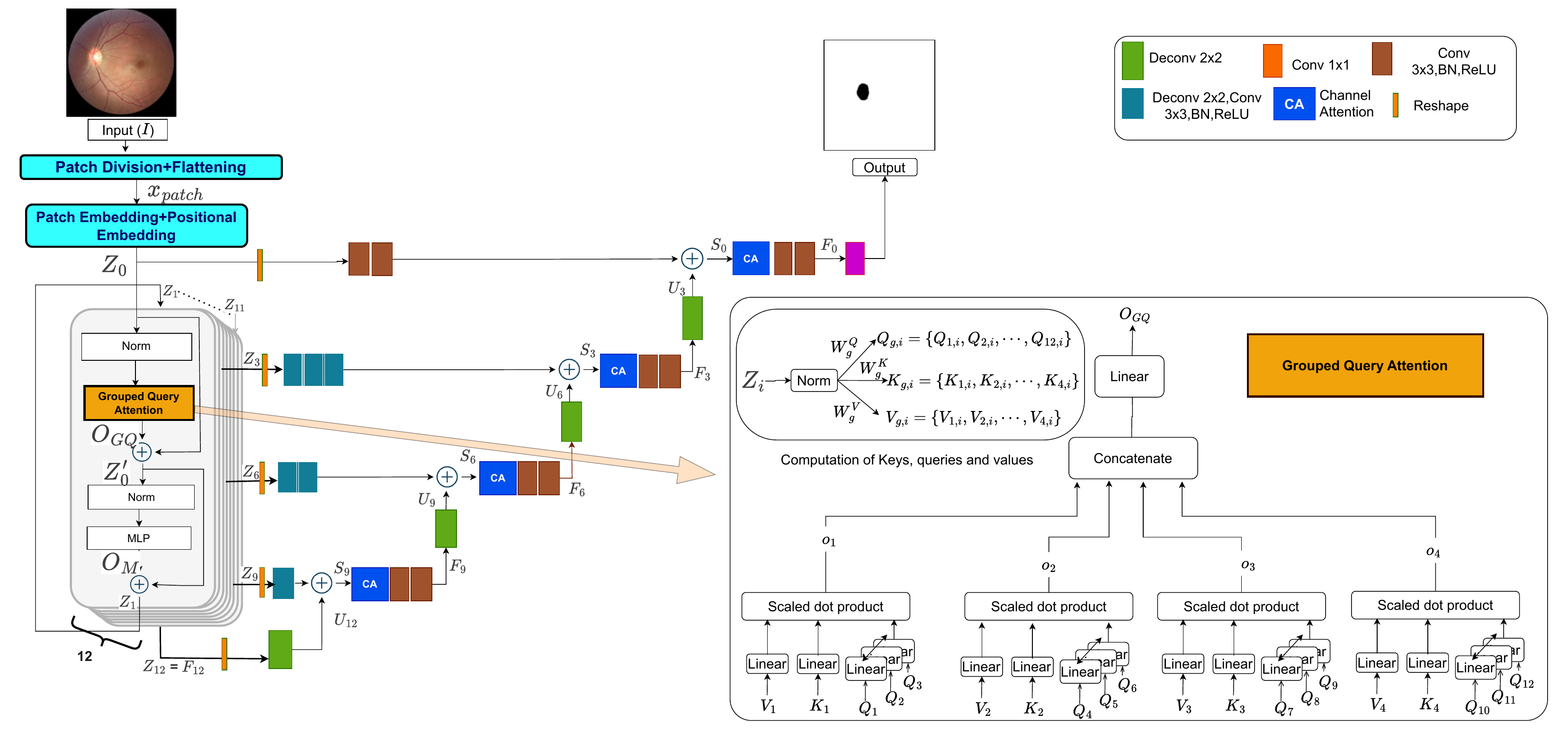}
    \caption{(Left) Block diagram of the proposed model OptiModNet. (Right) Details of the grouped query attention module.}
    \label{fig: blockdia}
\end{figure*}

The proposed methodology, depicted in Fig.~\ref{fig: blockdia}, addresses the challenges of semantic segmentation by integrating transformer-based encoders with convolutional decoders. This architecture incorporates two novel components to enhance both efficiency and accuracy. Firstly, Grouped Query Attention replaces the conventional multihead self-attention mechanism within the encoder, significantly reducing computational complexity without compromising performance on high-resolution medical images. Secondly, Channel Attention is employed in the decoder to adaptively emphasize salient feature channels hence attenuating less informative ones, this increasing segmentation accuracy.

\textit{Patch and Positional Embedding}:
Let the input image be {$\mathbf{I} \in \mathbb{R}^{H \times W \times C}$}, where $H$ and $W$ denote the height and width of the image, and $C$ represents the number of input channels. The input image is divided into flattened, uniform, non-overlapping patches {$\mathbf{I}_{patch} \in \mathbb{R}^{N \times (P^2 \cdot C)}$,} where $P \times P$ is the resolution of each patch and $N = \frac{H \cdot W}{P^2}$ is the number of patches i.e., the length of the sequence.

Each patch is projected into a embedding space of dimension $K$ using a learnable linear layer. To conserve spatial characteristics, a learnable $1D$ positional embedding $\mathbf{E}_{\text{pos}} \in \mathbb{R}^{N \times K}$ is added to the projected patch embedding $\mathbf{E}_{\text{patch}} \in \mathbb{R}^{(P^2 \cdot C) \times K}$. This ensures that the model can leverage spatial relationships across patches. The final embedding, $\mathbf{E}_{f}$, is computed as:

\begin{equation}
    \mathbf{E}_{f} = [\mathbf{x}_{patch}^{1} \cdot \mathbf{E}_{\text{patch}}; \mathbf{x}_{patch}^{2} \cdot \mathbf{E}_{\text{patch}}; \dots; \mathbf{x}_{patch}^{N} \cdot \mathbf{E}_{\text{patch}}] + \mathbf{E}_{\text{pos}},
\end{equation}
 
 where, $\mathbf{E}_{f} \in \mathbb{R}^{N \times K}$ is the output of the combined patch and positional embedding. It is sent as the input to the transformer encoder. 

\subsection{\textbf{Transformer Encoder}}
%
The encoder employs multiple stacked transformer blocks to effectively learn long-range interactions and contextual information from the sequence of patch and positional embeddings. Each transformer block consists of a grouped query attention mechanism followed by a feedforward network (MLP) and residual connections to ensure stable training. 

In standard transformer models, Multi-head Self-Attention (MSA) computes interactions between all input tokens. This can be expensive, especially for high-resolution inputs. To optimize this, we use Grouped Query Attention (GQA). This method reduces the computational load by splitting the queries into $G$ groups, while associating each group with a single key-value pair, thereby streamlining the attention mechanism.

Given an input $\mathbf{E}_{i-1}$, we first normalize it and then compute the Queries ($\mathbf{Q}_{g,i}$), Keys ($\mathbf{K}_{g,i}$), and Values ($\mathbf{V}_{g,i}$) for each group $g \in \{1, 2, \dots, G\}$, as follows 
:

\begin{equation}
\begin{aligned}
    \mathbf{Q}_{g,i} &= \text{Norm}(\mathbf{E}_{i-1}) \times \mathbf{W}_g^Q, \\
    \mathbf{K}_{g,i} &= \text{Norm}(\mathbf{E}_{i-1}) \times \mathbf{W}_g^K, \\
    \mathbf{V}_{g,i} &= \text{Norm}(\mathbf{E}_{i-1}) \times \mathbf{W}_g^V 
\end{aligned}
\end{equation}

where $\mathbf{W}_g^Q \in \mathbb{R}^{D \times 3q_g}, \mathbf{W}_g^K \in \mathbb{R}^{D \times k_g}, \mathbf{W}_g^V \in \mathbb{R}^{D \times v_g}$ are trainable weight matrices, and $q_g$, $k_g$, and $v_g$ denote the number of queries, keys, and values per group, respectively. For example, if there are 12 queries, 4 keys, and 4 values distributed over 4 groups, each group will contain 3 queries, 1 key, and 1 value. Attention for each group is computed as:
\begin{align}
    \mathbf{A}_{g,i} = \text{softmax}\left(\frac{\mathbf{Q}_{g,i} \mathbf{K}_{g,i}^\top}{\sqrt{k_g}}\right),
    \mathbf{O}_{g,i} = \mathbf{A}_{g,i} \mathbf{V}_{g,i}.
\end{align}

Once the attention outputs for all groups are computed, they are concatenated and projected back into the original embedding space:
\begin{equation}
    \mathbf{O}_{GQ,i} = \text{Concat}(\mathbf{O}_{1,i}, \mathbf{O}_{2,i}, \dots, \mathbf{O}_{G,i}) \mathbf{W}^O,
\end{equation}
where $\mathbf{W}^O \in \mathbb{R}^{3q_g G \times D}$ is the output projection matrix. This method significantly reduces the computational complexity of attention while maintaining the ability to model global dependencies in high-resolution inputs.

After grouped query attention, residual connections are added to get
$\mathbf{E}_{i-1}^{'} = \mathbf{E}_{i-1} + \mathbf{O}_{GQ,i}.$
This result is normalized, and a feedforward network (MLP) is added that refines the output features. This network consists of two linear layers with a Gaussian Error Linear Units (GELU) activation function and a dropout for regularization:

\begin{equation}
    \mathbf{O}_{M,i} = \text{Dropout}(\text{GELU}(Norm(\mathbf{E}_{i-1}^{'}) \mathbf{W}_1 + \mathbf{b}_1)) \mathbf{W}_2 + \mathbf{b}_2
\end{equation}

where $\mathbf{W}_1 \in \mathbb{R}^{D \times D'}$ and $\mathbf{W}_2 \in \mathbb{R}^{D' \times D}$, with $D'$ being the intermediate dimension. Adding the residual connection gives:
$\mathbf{F}_i = \mathbf{E}_{i-1}^{'} + \mathbf{O}_{M,i}.$
The $i^{th}$ transformer block outputs an embeddings $\mathbf{F}_i$, which is passed to the next $(i+1)^{th}$ block (shown in Fig.~\ref{fig: blockdia}). This process continues till the last $T^{th}$ transformer block, producing a final set of embeddings $\{F_1, F_2, \cdots, F_{T}\}$. In the proposed model we consider $T$ = 12.
%

\subsection{\textbf{ Convolutional Decoder}}
The decoder incrementally upsamples the feature maps while incorporating multi-resolution skip connections. Furthermore, the key contribution of the proposed architecture in this component is the incorporation of a channel attention layer. 

The output of the transformer encoder consists of features from various encoder resolutions, we extract a sequence representation $\mathbf{F}_i$, corresponding to specific layers $i \in \{12, 9, 6,3\}$. Each $\mathbf{F}_i$ is a 1D sequence with dimensions $\frac{H \times W}{P^2} \times C$, where $H$ and $W$ are the input image's spatial dimensions, $P$ is the patch size, and $C$ is the embedding dimension. 

We include the encoder's outputs as skip connection within the decoder. To match the spatial dimension they are reshaped into 2D feature maps as the following,
$\mathbf{F}_i = \text{Reshape}(\mathbf{F}_i)$, $\mathbf{F}_i \in \mathbb{R}^{\frac{H}{P} \times \frac{W}{P} \times C}$. The reshaped features are passed through  $2 \times 2$ deconvolutional layers and $3 \times 3$ convolutional layers followed by batch normalization (BN) and ReLU activation:
\begin{equation}
    \mathbf{C}_i = \text{ReLu}(\text{BN}(\text{Conv2D}(\text{Conv2DTranspose}(\mathbf{F}_i))))
\end{equation}
This ensures that the features are better aligned for subsequent processing in the decoder.
\subsubsection{Hierarchical Decoder Workflow}
The decoder operates in a stepwise manner, starting from the highest-resolution feature map $\mathbf{F}_{12}$  and progressively reconstructing the input resolution. Each step of the decoder involves four key operations: Upsampling, Skip Connection Integration, Channel Attention, and Convolutional Refinement. 

\textbf{Upsampling with Deconvolution:} At each stage, the spatial resolution of the feature map is doubled using a $2 \times 2$ transposed convolution:
    $\mathbf{U}_i = \text{Conv2DTranspose}(\mathbf{F}_i), \quad \mathbf{U}_i \in \mathbb{R}^{2H' \times 2W' \times C}.$
This operation enlarges the feature map while preserving its learned semantic information.

\textbf{Integration of Multi-Resolution Features}: The upsampled feature map $\mathbf{U}_i$ is combined with the corresponding encoder feature $\mathbf{F}_{i-3}$ through an element-wise addition:
$
    \mathbf{S}_{i-3} = \mathbf{U}_i + \mathbf{F}_{i-3},
$
This step enriches the upsampled features with contextual information from the encoder.

\textbf{Channel Attention Mechanism} (CA): The combined feature map $\mathbf{S}_{i-3}$ is passed through a channel attention module to recalibrate feature importance. The channel attention weights are computed as follows, first, a Global Average Pooling (GAP) operation is applied to $\mathbf{S}_{i-3}$ to reduce its spatial dimensions: $\mathbf{C}_{\text{avg}} = \text{GlobalAvgPool}(\mathbf{S}_{i-3}), \quad \mathbf{C}_{\text{avg}} \in \mathbb{R}^{C}$. The pooled vector $\mathbf{C}_{\text{avg}}$ is then passed through two fully connected layers with ReLU and sigmoid activations to compute the attention weights, such as $\mathbf{C}_{\text{attn}} = \sigma\left( \text{Dense}\left(\text{ReLU}\left(\text{Dense}(\mathbf{C}_{\text{avg}})\right)\right)\right)$, $\mathbf{C}_{\text{attn}} \in \mathbb{R}^{C}$.
The weights $\mathbf{C}_{\text{attn}}$ are reshaped and applied to $\mathbf{S}_{i-3}$ through element-wise multiplication as $\mathbf{S}_{i-3}' = \mathbf{S}_{i-3} \odot \text{Reshape}(\mathbf{C}_{\text{attn}})$. This recalibration emphasizes informative channels while reducing the impact of less relevant ones.

\textbf{Convolutional Refinement}: The channel-attended feature map $\mathbf{S}_{i-3}'$ is processed through a sequence of convolutional layers to enhance its spatial and semantic representation: $\mathbf{F}_{i-3} = \text{ReLU}(\text{BN}(\text{Conv2D}(\mathbf{S}_{i-3}')))$. The refined feature map $\mathbf{F}_{i-3}$ serves as the input for the next decoder stage, continuing the upsampling process.

\textbf{Final Decoding}: At the final stage of the decoder, the reconstructed feature map $\mathbf{F}_{0}$ is passed through a $1 \times 1$ convolution to reduce the number of channels to the target classes. The final pixel-wise outputs are obtained by applying a sigmoid activation function:$
\mathbf{Y} = \sigma(\text{Conv2D}(\mathbf{F}_{0})), \quad \mathbf{Y} \in \mathbb{R}^{H \times W}$. Here, $\mathbf{Y}$ represents the segmentation map, with pixel values indicating the predicted probabilities for each class.

\subsection{\textbf{Aggregated Pyramid Loss}}
%
To effectively optimize both region-level overlap and pixel-wise classification accuracy, we adopt a hybrid loss function termed as aggregated pyramid loss that combines the soft Dice loss and Binary Cross-Entropy (BCE) loss. This combined loss is computed in a pixel-wise manner and is defined as:
\begin{equation}
\small
\begin{aligned}
\mathcal{L}(\text{GT}, \text{Y}) =\ & 1 - 
\frac{2 \sum_{i=1}^{M} \text{GT}_i \cdot \text{Y}_i + \epsilon}
{\sum_{i=1}^{M} \text{GT}_i + \sum_{i=1}^{M} \text{Y}_i + \epsilon} \\
& - \frac{1}{M} \sum_{i=1}^{M} \Big[ 
\text{GT}_i \cdot \log(\text{Y}_i) +
(1 - \text{GT}_i) \cdot \log(1 - \text{Y}_i) 
\Big]
\end{aligned}
\end{equation}

where $\text{GT}_i$ denotes the ground truth label,
$\text{Y}_i$ denotes the predicted probability from Final Decoding,
$M$ is the total number of pixels ($i$),
$\epsilon$ is a small constant for numerical stability.

Given the presence of multiple output stages in the network, we compute the loss at each stage and aggregate them using a weighted sum - hence the name Aggregated Pyramid Loss. Let $\mathcal{L}_k$ denote the combined loss at the $k$-th output, and $\lambda_k$ be the corresponding weight. The total loss used for optimization is, $
\mathcal{L}_{\text{total}} = \sum_{k=1}^{K} \lambda_k \cdot \mathcal{L}_k,    
$ where \(K\) is the number of output branches (e.g., intermediate and final outputs), and \(\lambda_k\) controls the contribution of each output to the total loss. {In our implementation we have 3 intermediate output for which we set $\lambda = \{0.2, 0.3, 0.5\}$ and for the final output $\lambda = 1$.} The intuition is to ensure supervision at multiple stages while prioritizing the final prediction. An ablation study corresponding to the different values of $\lambda$ are given the Sec.~\ref{sec: ablation}. 
%
\subsection{Implementation details}
This section deals with the detailed implementation parameters and hyperparameters utilized for our model.
We resized all the images to 256×256 pixels and normalized to the range [0, 1]. Each image was divided into 16×16 pixel patches to preserve spatial features essential for segmentation. The corresponding masks were resized, normalized.
The model processes input images with 3 RGB channels and follows a 12-layer encoder-decoder structure. It employs a hidden dimension of 128 for feature representation and an MLP dimension of 32 for the multi-layer perceptron layers. The transformer architecture includes 12 query heads and 4 key-value heads for multi-head attention, ensuring effective feature extraction and segmentation performance. The model was trained using the proposed Aggregated Pyramid Loss, which supervises each intermediate and final output with a combined BCE and Dice loss. We employed the Adam optimizer with a learning rate of $1 \times 10^{-3}$. Training was conducted for 50 epochs with a batch size of 8.

\section{Experiments}
\label{sec: experiments}
In this section, we conduct quantitative experimental analyses along with an ablation study. We compare with some benchmark methods
on two standard segmentation metrics, Dice Coefficient(Dice) and Intersection Over Union(IOU), on two optic disc (OD) and optic cup (OC) segmentation benchmark datasets - REFUGE2 (Disc and cup)~\cite{orlando2020refuge} and ORIGA~\cite{zhang2010origa}. Table~\ref{tab: comparison_refuge} and Table.~\ref{tab: comparison_orega} presents the comprehensive comparison of segmentation methods across these datasets.

\textit{Dataset details} - The REFUGE2 dataset contains a total of 2,000 RGB fundus images, each accompanied by binary masks representing the optic disc and cup areas. The dataset is used with the standard splits into training, validation, and test sets. The ORIGA dataset contains 650 retinal fundus images, including 168 from glaucomatous patients. It provides precise optic disc (OD) and optic cup (OC) boundary annotations, along with cup-to-disc ratio (CDR) values, making it suitable for segmentation-based glaucoma analysis. The images typically have low brightness and noticeable noise, posing additional challenges for automated segmentation methods.
\begin{table}[t]
\renewcommand{\arraystretch}{1.1}
\centering
\caption{Comparative study using Dice and IoU on REFUGE2 dataset.}
\label{tab: comparison_refuge}
\begin{tabular}{p{3.5cm}cccc}
\toprule
\multirow{2}{*}{Method} & \multicolumn{2}{c}{REFUGE2-Disc} & \multicolumn{2}{c}{REFUGE2-Cup} \\ \cline{2-5}
 & Dice & IoU & Dice & IoU \\ \hline
ResUNet~\cite{diakogiannis2020resunet} & 92.90 & 85.50 & 80.10 & 72.30 \\
BEAL~\cite{wang2019boundary} & 93.70 & 86.10 & 83.50 & 74.10 \\
TransBTS~\cite{wenxuan2021transbts} & 92.30 & 85.80 & 83.70 & 75.90 \\
SwinBTS~\cite{jiang2022swinbts} & 95.20 & 87.70 & 86.30 & 75.70 \\
BAT~\cite{wang2021boundary} & 92.30 & 85.80 & 82.00 & 73.20 \\
nnUNet~\cite{isensee2021nnu} & 94.50 & 87.70 & 83.60 & 75.40 \\
TransUNet~\cite{chen2021transunet} & 95.00 & 87.70 & 85.40 & 77.20 \\
UNetr~\cite{hatamizadeh2022unetr} & 94.60 & 87.80 & 83.50 & 75.70 \\
Swin-UNetr~\cite{hatamizadeh2021swin} & 95.30 & 87.90 & 85.40 & 77.30 \\
EnsemDiff~\cite{wolleb2022diffusion} & 92.60 & 85.20 & 83.00 & 73.40 \\
SegDiff~\cite{amit2021segdiff} & 92.10 & 85.20 & 83.00 & 73.40 \\
MedsegDiff~\cite{wu2024medsegdiff}+TransUNet & 91.80 & 84.50 & 82.00 & 72.60 \\
MedSegDiff-V2~\cite{wu2024medsegdiffv2} & 96.70 & 88.90 & 87.90 & 80.30 \\
\textbf{OptiModNet (Ours)} & \textbf{99.45} & \textbf{98.16} & \textbf{93.23} & \textbf{89.41} \\

\hline
\end{tabular}
\end{table}
\begin{table}[!h]
\centering
\caption{Segmentation performance on ORIGA dataset for 3 class classification, optic cup, optic disc and background.}
\label{tab: comparison_orega}
\begin{tabular}{lcc}
\toprule
\textbf{Model} & \textbf{IoU} & \textbf{Dice} \\
\midrule
U-Net~\cite{ronneberger2015u} & 91.58 & 92.76 \\
Attention U-Net~\cite{oktay2018attention} & 92.39 & 95.39 \\
DeepLabV3+~\cite{chen2018encoder} & 95.10 & 96.29 \\
TransUNet~\cite{chen2021transunet} & 96.29 & 96.29 \\
Swin-UNet~\cite{liu2021swin} & 96.00 & 98.79 \\
Wang et al.~\cite{wang2025boundary} & 97.60 & 98.79 \\
OptiModNet(Ours) & \textbf{98.92} & \textbf{99.31} \\
\bottomrule
\end{tabular}
\end{table}

\subsection{Quantitative Experiments}
In Table~\ref{tab: comparison_refuge} we can observe, the proposed OptiModNet model outperforms all other methods on the REFUGE2 dataset for both optic cup and disc segmentation.  Our proposed approach, OptiModNet, achieves the highest performance across all metrics. Specifically, for the disc segmentation task, OptiModNet attains a Dice score of 99.45\% and an IoU of 98.16\%, surpassing the next-best method (MedSegDiff-V2) by 2.75\% and 2.26\%, respectively. For the cup segmentation task, OptiModNet attains a Dice score of 93.23\% and an IoU of 89.41\%, outperforming the second-best method MedSegDiff-V2 by 2.23\% and 1.11\%, respectively.

Table.~\ref{tab: comparison_orega} reports the segmentation performance of various methods on the ORIGA dataset for the three-class classification task (optic cup, optic disc, and background). Our proposed method, OptiModNet, achieves the highest scores across both metrics, with an IoU of 98.92\% and a Dice score of 99.31\%. Compared to the next-best method by Wang et al., OptiModNet improves IoU by 1.32\% and Dice by 0.52\%.

The improvement margin is more pronounced compared to the classical architectures such as U-Net and Attention U-Net, where the proposed method demonstrates an IoU gain of 7.34\% and 6.53\%, respectively, and a Dice gain of 6.55\% and 6.92\%, respectively. These results indicate the efficacy of OptiModNet compared to modern transformer-based models like Swin-UNet and TransUNet as well as traditional convolutional baselines. The superior performance on ORIGA, especially in IoU, shows the model’s precise boundary delineation capabilities, which are essential for accurate localization of optic disc and cup regions in clinical settings.

\begin{figure*}[!h]
    \centering
    \includegraphics[width=\linewidth]{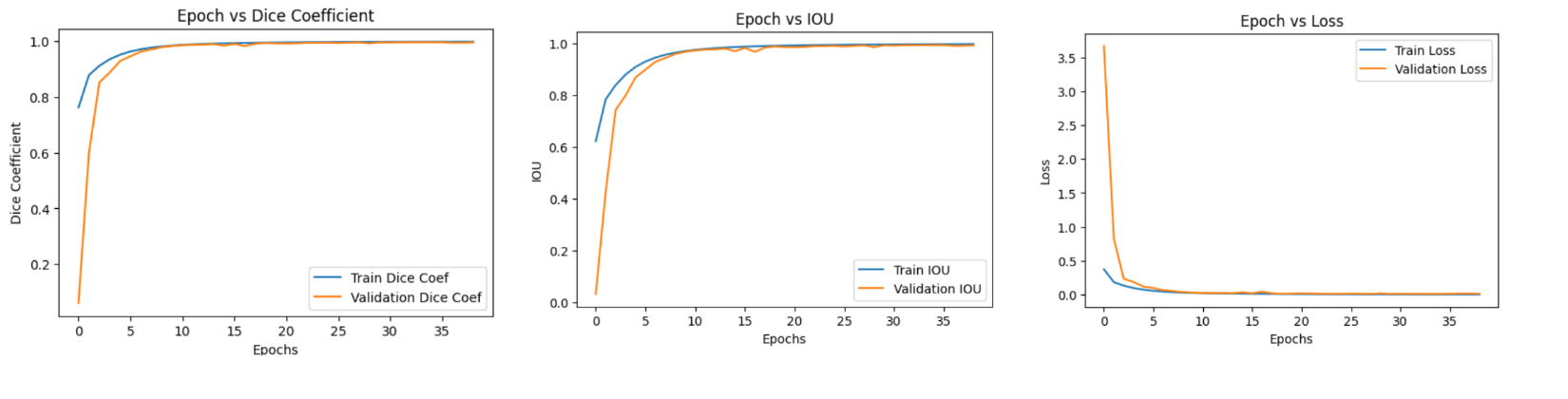}
    \caption{Training and validation curves for Dice, IoU, and loss on the REFUGE2 (optic disc) dataset show increasing metric values over epochs, and decreasing loss indicating convergence.}
    \label{fig:loss_graphs}
\end{figure*}

Fig.~\ref{fig:loss_graphs} shows the performance metrics and loss of our method over 50 epochs on the REFUGE2 (optic disc) dataset. The Dice and IoU curves exhibit rapid improvement in both training and validation, reflecting strong alignment with ground truth.

\subsection{Qualitative Experiments}\label{sec:quan}
\begin{figure}[!h]
    \centering
    \begin{minipage}[b]{0.45\linewidth}
        \centering
        \includegraphics[width=\linewidth]{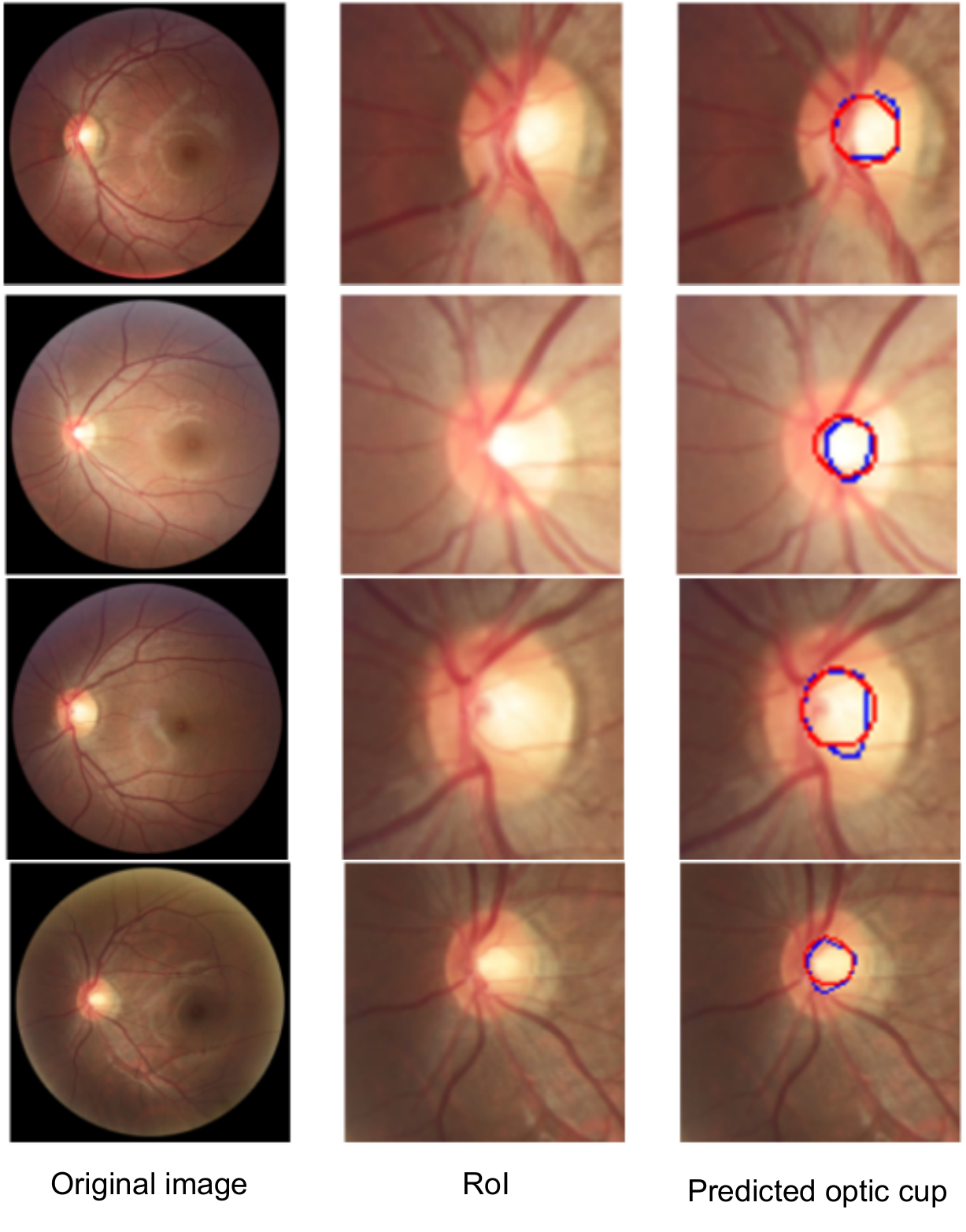}
        \caption{Qualitative results of optic cup segmentation on the REFUGE2 testset.}
        \label{fig:qualitative1}
    \end{minipage}
    \hfill
    \begin{minipage}[b]{0.45\linewidth}
        \centering
        \includegraphics[width=0.97\linewidth]{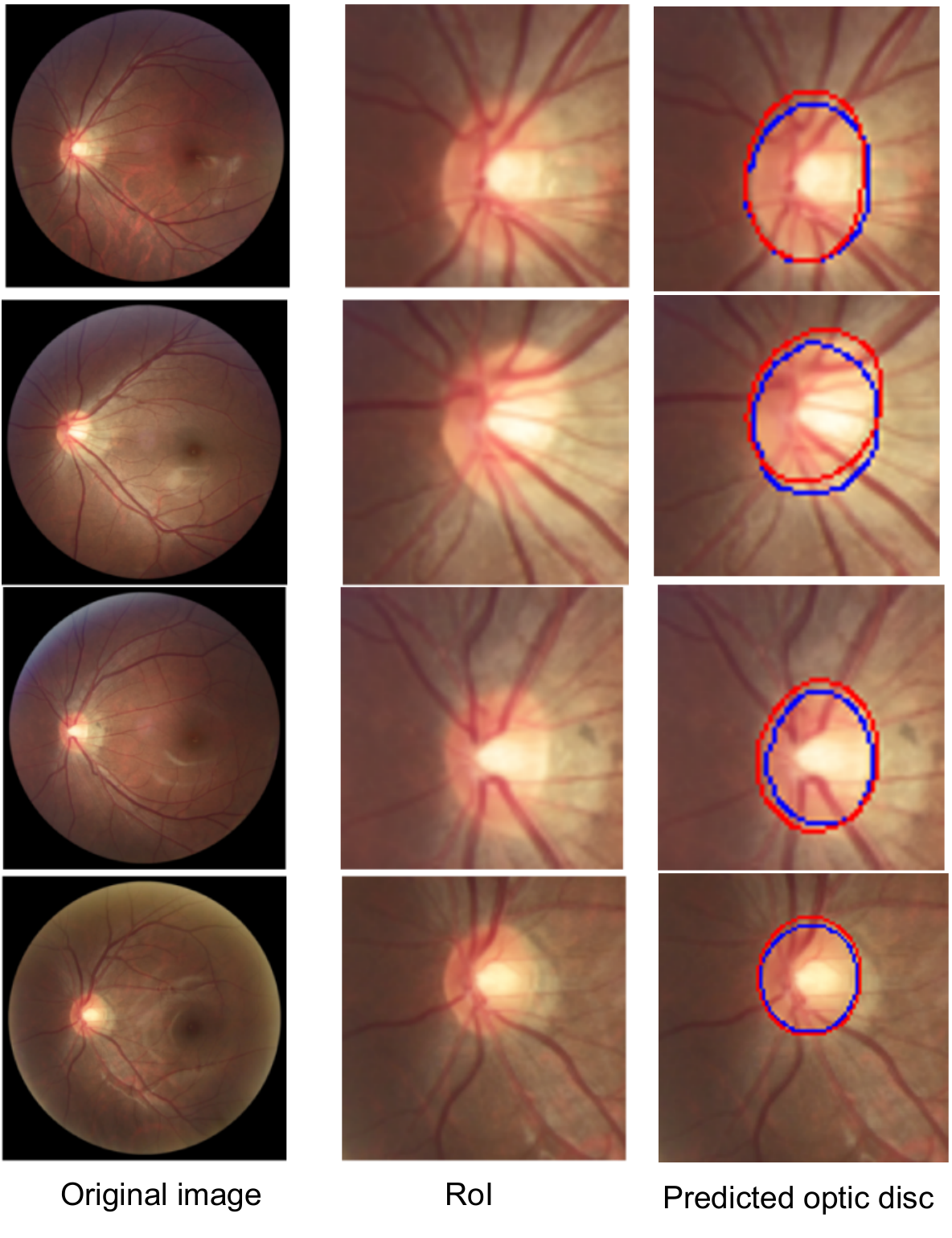}
        \caption{Qualitative results of optic disc segmentation on the REFUGE2 testset.}
        \label{fig:qualitative2}
    \end{minipage}
\end{figure}

Fig.~\ref{fig:qualitative1}, Fig.~\ref{fig:qualitative2} shows our results of optic cup and disc segmentations on the REFUGE2 dataset. In most examples, the predicted masks align closely with the ground truth boundaries, indicating better localization and shape preservation of the optic disc. In general across the results the predictions exhibit smooth and well-defined contours, with only minor deviations near the boundaries, including cases where the disc margin is less distinct due to illumination variation or vessel occlusion. It is observable that these visual observations are consistent with the Dice and IoU scores in Table.~\ref{tab: comparison_refuge}, indicating that the qualitative performance aligns well with the reported quantitative results.

\subsection{Parameters}\label{sec:parameters}
Table~\ref{tab:model_comparison} presents a comparison of various models based on their parameter counts and computational complexity (measured in Gflops). OptiModNet stands out with the lowest parameter count (1.9M), significantly smaller than others (96\% parameters efficient compared to MedSegdiff v2) highlighting its efficiency while potentially maintaining competitive performance. A related reduction plot is shown in Figure~\ref{fig:reduction}, highlighting that the proposed model retains nearly 96\% fewer parameters compared to the highest-performing state-of-the-art models. In addition to parameter efficiency, the proposed OptiModNet achieves the second-best performance while requiring only 0.2\% of the GFlops used by MedSegDiff-V2.

\begin{minipage}[b]{0.4\textwidth}
\centering
\captionof{table}{Comparison of models on Params and Gflops}
\resizebox{1\linewidth}{!}{
\begin{tabular}{l|c|c}
\hline
Model & Params (M) & Gflops \\
\hline
Swin-UNet & 21 & 34 \\
EnsemDiff & 23 & 2203 \\
SegDiff & 23 & 2399 \\
MedSegDiff & 25 & 1770 \\
MedSegDiff-V2 & 46 & 983 \\
OptiModNet  &  \textbf{1.93} & \textbf{3.73} \\
\hline
\end{tabular}
}
\label{tab:model_comparison}
\end{minipage}
\begin{minipage}[t]{0.5\textwidth}
\centering
\includegraphics[width=0.9\linewidth, trim=0 0 0 0, clip]{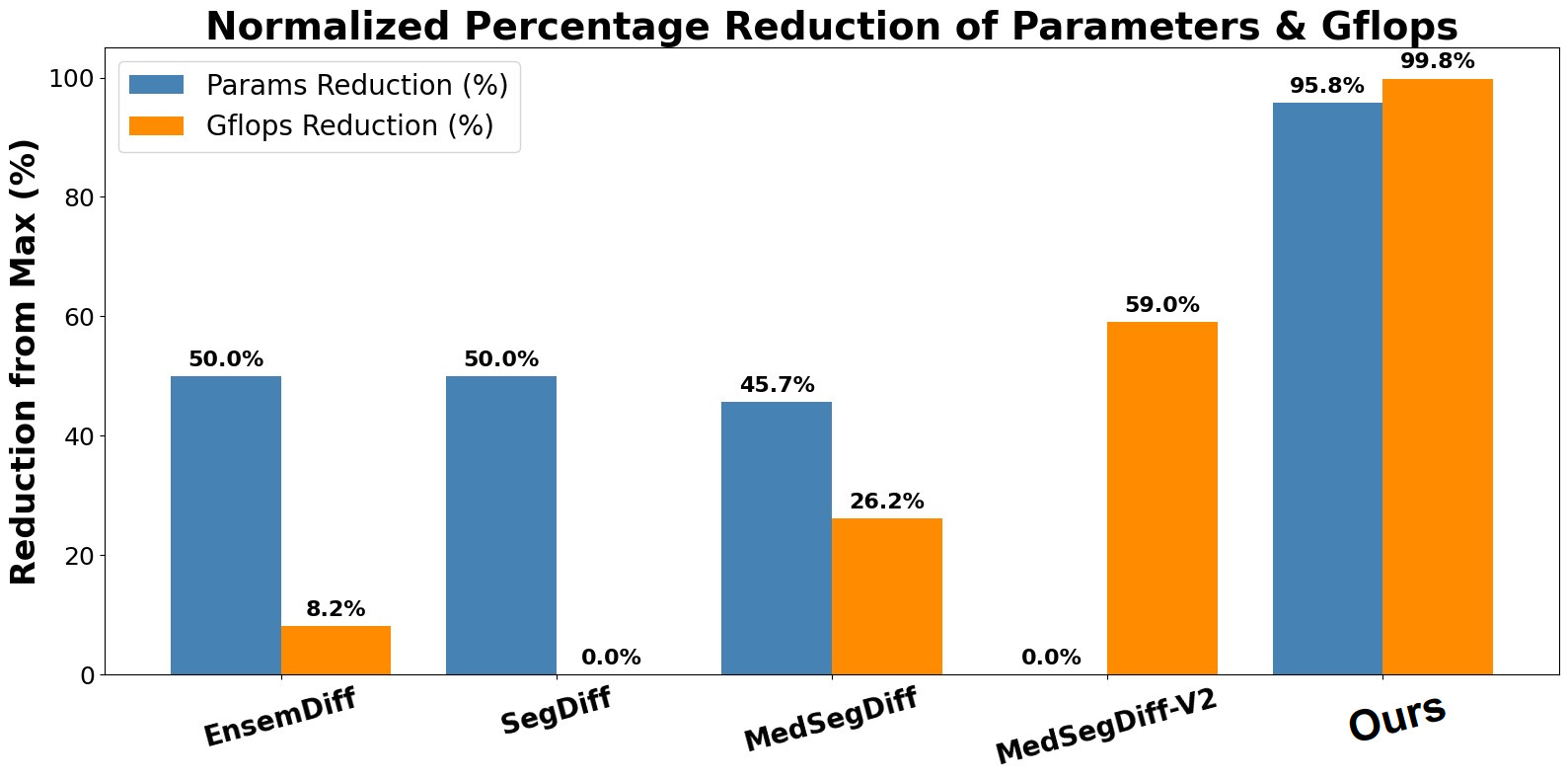}
\captionof{figure}{Reduction plot}
\label{fig:reduction}
\end{minipage}
\begin{figure}[!h]
    \centering
    \includegraphics[width=0.7\linewidth]{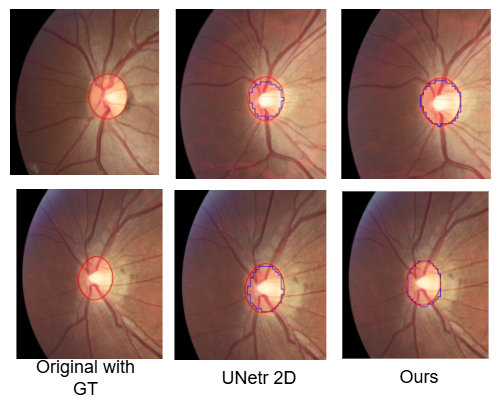}
    \caption{Comparative study of our result with that of UNETR 2D model on REFUGE-2 disc dataset. For better visual comparability, we have extracted the boundary of the mask and overlaid on the input image. It can be observed that the boundary of our predicted mask (blue) aligns comparatively better with the ground truth (red).}
    \label{fig:ablation}  
\end{figure}
\subsection{Ablation Study}
\label{sec: ablation}
To understand the effect of each attention module an ablation study is presented in Table.~\ref{tab:ablation}. The incremental improvement in scores confirms that each attention module contributes meaningfully to performance improvement. Table.~\ref{tab:lambda_ablation} presents the ablation over different $\lambda$ values corresponding to the proposed loss function on REFUGE2-disc dataset. A visual ablation presented in Figure.~\ref{fig:ablation} compares our performance with the UNetr 2D exhibiting comparatively  that OptiModNet shows improved boundary adherence and captures better fine details.
\begin{table}[!h]
\centering
\begin{minipage}[t]{0.6\textwidth}
\centering
\scriptsize
\caption{Ablation study: Comparison of OptiModNet with various modifications on the baseline UNETR 2D. 
Note. CA - Channel Attention and GQA - Grouped Query Attention.}
\label{tab:ablation}
\begin{tabular}{@{}p{2.5cm}cccc@{}}
    \toprule
    Configuration & \multicolumn{2}{c}{REFUGE2-Disc} & \multicolumn{2}{c}{REFUGE2-Cup} \\
    \cmidrule(lr){2-3} \cmidrule(lr){4-5}
                  & Dice (\%)  & IoU (\%)  & Dice (\%)  & IoU (\%) \\
    \midrule
    UNETR 2D                  & 94.21 & 93.83 & 85.27 & 80.80 \\
    UNETR 2D + CA              & 95.60 & 94.28 & 86.23 & 81.30 \\
    UNETR 2D + GQA             & 97.23 & 96.75 & 89.89 & 87.56 \\
    UNETR 2D + CA + GQA (Ours) & \textbf{99.45} & \textbf{98.16} & \textbf{93.23} & \textbf{89.41} \\
    \bottomrule
\end{tabular}
\end{minipage}
\hfill
\begin{minipage}[t]{0.38\textwidth}
\centering
\scriptsize
\caption{Ablation study of our approach with different $\lambda$ values on the REFUGE 2-disc dataset.}
\label{tab:lambda_ablation}
\begin{tabular}{ccccc|p{1cm}p{1cm}}
\toprule
$\lambda_1$ & $\lambda_2$ & $\lambda_3$ & $\lambda_4$ & & Dice (\%) & IoU (\%) \\
\midrule
1    & 1    & 1    & 1    & & 99.03 & 97.56 \\
0.5  & 0.5  & 0.5  & 0.5  & & 99.41 & 98.02 \\
0.25 & 0.25 & 0.25 & 0.25 & & 99.38 & 98.09 \\
0.2  & 0.3  & 0.5  & 1    & & \textbf{99.45} & \textbf{98.16} \\
\bottomrule
\end{tabular}
\end{minipage}
\end{table}

\section{Conclusion and Future Work}
\label{sec: conclusion}

In this paper, we propose OptiModNet, a novel framework for optic disc and cup segmentation that extends the UNetR (UNet with Transformers) architecture by integrating Grouped Query Attention and a newly introduced Aggregated Pyramid Loss function. OptiModNet demonstrates state-of-the-art performance on the REFUGE2 and ORIGA datasets. On REFUGE2, it achieves Dice scores of 99.45\% for disc and 93.23\% for cup, surpassing existing approaches by over 2.5\%. On ORIGA, it further attains the highest reported segmentation accuracy, despite requiring nearly 96\% fewer parameters than the most complex state-of-the-art model. These results highlight the effectiveness and efficiency of OptiModNet in advancing OC/OD segmentation.

\section{Acknowledgements}
We thank Dr. Swalpa Kumar Roy for his insightful suggestions and guidance. This work was partially supported by the Anusandhan National Research Foundation (ANRF) through the Prime Minister Early Career Research Grant. 
\bibliographystyle{splncs04}
\bibliography{references}

@article{wang2025boundary,
  title={Boundary-Aware Transformer for Optic Cup and Disc Segmentation in Fundus Images},
  author={Wang, Soohyun and Kim, Byoungkug and Eom, Doo-Seop},
  journal={Applied Sciences},
  volume={15},
  number={9},
  pages={5165},
  year={2025},
  publisher={MDPI}
}

@inproceedings{zhang2010origa,
  title={Origa-light: An online retinal fundus image database for glaucoma analysis and research},
  author={Zhang, Zhuo and Yin, Feng Shou and Liu, Jiang and Wong, Wing Kee and Tan, Ngan Meng and Lee, Beng Hai and Cheng, Jun and Wong, Tien Yin},
  booktitle={2010 Annual international conference of the IEEE engineering in medicine and biology},
  pages={3065--3068},
  year={2010},
  organization={IEEE}
}

@article{jiang2019jointrcnn,
  title={JointRCNN: a region-based convolutional neural network for optic disc and cup segmentation},
  author={Jiang, Yuming and Duan, Lixin and Cheng, Jun and Gu, Zaiwang and Xia, Hu and Fu, Huazhu and Li, Changsheng and Liu, Jiang},
  journal={IEEE Transactions on Biomedical Engineering},
  volume={67},
  number={2},
  pages={335--343},
  year={2019},
  publisher={IEEE}
}

@article{sevastopolsky2017optic,
  title={Optic disc and cup segmentation methods for glaucoma detection with modification of U-Net convolutional neural network},
  author={Sevastopolsky, Artem},
  journal={Pattern Recognition and Image Analysis},
  volume={27},
  number={3},
  pages={618--624},
  year={2017},
  publisher={Springer}
}

@article{yu2019robust,
  title={Robust optic disc and cup segmentation with deep learning for glaucoma detection},
  author={Yu, Shuang and Xiao, Di and Frost, Shaun and Kanagasingam, Yogesan},
  journal={Computerized Medical Imaging and Graphics},
  volume={74},
  pages={61--71},
  year={2019},
  publisher={Elsevier}
}

@article{liu2019spatial,
  title={A spatial-aware joint optic disc and cup segmentation method},
  author={Liu, Qing and Hong, Xiaopeng and Li, Shuo and Chen, Zailiang and Zhao, Guoying and Zou, Beiji},
  journal={Neurocomputing},
  volume={359},
  pages={285--297},
  year={2019},
  publisher={Elsevier}
}

@article{yi2023c2ftfnet,
  title={C2FTFNet: Coarse-to-fine transformer network for joint optic disc and cup segmentation},
  author={Yi, Yugen and Jiang, Yan and Zhou, Bin and Zhang, Ningyi and Dai, Jiangyan and Huang, Xin and Zeng, Qinqin and Zhou, Wei},
  journal={Computers in Biology and Medicine},
  volume={164},
  pages={107215},
  year={2023},
  publisher={Elsevier}
}

@article{zhao2021application,
  title={Application of an attention u-net incorporating transfer learning for optic disc and cup segmentation},
  author={Zhao, Xiaoye and Wang, Shengying and Zhao, Jing and Wei, Haicheng and Xiao, Mingxia and Ta, Na},
  journal={Signal, Image and Video Processing},
  volume={15},
  number={5},
  pages={913--921},
  year={2021},
  publisher={Springer}
}

@article{yan2024mrsnet,
  title={MRSNet: Joint consistent optic disc and cup segmentation based on large kernel residual convolutional attention and self-attention},
  author={Yan, Shiliang and Pan, Xiaoqin and Wang, Yinling},
  journal={Digital Signal Processing},
  volume={145},
  pages={104308},
  year={2024},
  publisher={Elsevier}
}

@article{pachade2021nenet,
  title={NENet: Nested EfficientNet and adversarial learning for joint optic disc and cup segmentation},
  author={Pachade, Samiksha and Porwal, Prasanna and Kokare, Manesh and Giancardo, Luca and Meriaudeau, Fabrice},
  journal={Medical Image Analysis},
  volume={74},
  pages={102253},
  year={2021},
  publisher={Elsevier}
}

@inproceedings{chen2018encoder,
  title={Encoder-decoder with atrous separable convolution for semantic image segmentation},
  author={Chen, Liang-Chieh and Zhu, Yukun and Papandreou, George and Schroff, Florian and Adam, Hartwig},
  booktitle={Proceedings of the European conference on computer vision (ECCV)},
  pages={801--818},
  year={2018}
}

@article{oktay2018attention,
  title={Attention u-net: Learning where to look for the pancreas},
  author={Oktay, Ozan and Schlemper, Jo and Folgoc, Loic Le and Lee, Matthew and Heinrich, Mattias and Misawa, Kazunari and Mori, Kensaku and McDonagh, Steven and Hammerla, Nils Y and Kainz, Bernhard and others},
  journal={arXiv preprint arXiv:1804.03999},
  year={2018}
}

@article{dosovitskiy2020image,
  title={An image is worth 16x16 words},
  author={Dosovitskiy, Alexey and Beyer, Lucas and Kolesnikov, Alexander and Weissenborn, Dirk and Zhai, Xiaohua and Unterthiner, Thomas and Dehghani, Mostafa and Minderer, Matthias and Heigold, Georg and Gelly, Sylvain and others},
  journal={arXiv preprint arXiv:2010.11929},
  volume={7},
  year={2020}
}

@inproceedings{liu2021swin,
  title={Swin transformer: Hierarchical vision transformer using shifted windows},
  author={Liu, Ze and Lin, Yutong and Cao, Yue and Hu, Han and Wei, Yixuan and Zhang, Zheng and Lin, Stephen and Guo, Baining},
  booktitle={Proceedings of the IEEE/CVF international conference on computer vision},
  pages={10012--10022},
  year={2021}
}

@inproceedings{ronneberger2015u,
  title={U-net: Convolutional networks for biomedical image segmentation},
  author={Ronneberger, Olaf and Fischer, Philipp and Brox, Thomas},
  booktitle={Medical image computing and computer-assisted intervention--MICCAI 2015: 18th international conference, Munich, Germany, October 5-9, 2015, proceedings, part III 18},
  pages={234--241},
  year={2015},
  organization={Springer}
}

@article{chen2021transunet,
  title={Transunet: Transformers make strong encoders for medical image segmentation},
  author={Chen, Jieneng and Lu, Yongyi and Yu, Qihang and Luo, Xiangde and Adeli, Ehsan and Wang, Yan and Lu, Le and Yuille, Alan L and Zhou, Yuyin},
  journal={arXiv preprint arXiv:2102.04306},
  year={2021}
}

@inproceedings{cao2022swin,
  title={Swin-unet: Unet-like pure transformer for medical image segmentation},
  author={Cao, Hu and Wang, Yueyue and Chen, Joy and Jiang, Dongsheng and Zhang, Xiaopeng and Tian, Qi and Wang, Manning},
  booktitle={European conference on computer vision},
  pages={205--218},
  year={2022},
  organization={Springer}
}

@inproceedings{hu2018squeeze,
  title={Squeeze-and-excitation networks},
  author={Hu, Jie and Shen, Li and Sun, Gang},
  booktitle={Proceedings of the IEEE conference on computer vision and pattern recognition},
  pages={7132--7141},
  year={2018}
}

@article{ainslie2023gqa,
  title={Gqa: Training generalized multi-query transformer models from multi-head checkpoints},
  author={Ainslie, Joshua and Lee-Thorp, James and de Jong, Michiel and Zemlyanskiy, Yury and Lebr{\'o}n, Federico and Sanghai, Sumit},
  journal={arXiv preprint arXiv:2305.13245},
  year={2023}
}

@inproceedings{hatamizadeh2022unetr,
  title={Unetr: Transformers for 3d medical image segmentation},
  author={Hatamizadeh, Ali and Tang, Yucheng and Nath, Vishwesh and Yang, Dong and Myronenko, Andriy and Landman, Bennett and Roth, Holger R and Xu, Daguang},
  booktitle={Proceedings of the IEEE/CVF winter conference on applications of computer vision},
  pages={574--584},
  year={2022}
}

@article{orlando2020refuge,
  title={Refuge challenge: A unified framework for evaluating automated methods for glaucoma assessment from fundus photographs},
  author={Orlando, Jos{\'e} Ignacio and Fu, Huazhu and Breda, Jo{\~a}o Barbosa and Van Keer, Karel and Bathula, Deepti R and Diaz-Pinto, Andr{\'e}s and Fang, Ruogu and Heng, Pheng-Ann and Kim, Jeyoung and Lee, JoonHo and others},
  journal={Medical image analysis},
  volume={59},
  pages={101570},
  year={2020},
  publisher={Elsevier}
}

@article{diakogiannis2020resunet,
  title={ResUNet-a: A deep learning framework for semantic segmentation of remotely sensed data},
  author={Diakogiannis, Foivos I and Waldner, Fran{\c{c}}ois and Caccetta, Peter and Wu, Chen},
  journal={ISPRS Journal of Photogrammetry and Remote Sensing},
  volume={162},
  pages={94--114},
  year={2020},
  publisher={Elsevier}
}

@inproceedings{wolleb2022diffusion,
  title={Diffusion models for implicit image segmentation ensembles},
  author={Wolleb, Julia and Sandk{\"u}hler, Robin and Bieder, Florentin and Valmaggia, Philippe and Cattin, Philippe C},
  booktitle={International Conference on Medical Imaging with Deep Learning},
  pages={1336--1348},
  year={2022},
  organization={PMLR}
}

@article{amit2021segdiff,
  title={Segdiff: Image segmentation with diffusion probabilistic models},
  author={Amit, Tomer and Shaharbany, Tal and Nachmani, Eliya and Wolf, Lior},
  journal={arXiv preprint arXiv:2112.00390},
  year={2021}
}

@inproceedings{wu2024medsegdiff,
  title={Medsegdiff: Medical image segmentation with diffusion probabilistic model},
  author={Wu, Junde and Fu, Rao and Fang, Huihui and Zhang, Yu and Yang, Yehui and Xiong, Haoyi and Liu, Huiying and Xu, Yanwu},
  booktitle={Medical Imaging with Deep Learning},
  pages={1623--1639},
  year={2024},
  organization={PMLR}
}

@inproceedings{wu2024medsegdiffv2,
  title={Medsegdiff-v2: Diffusion-based medical image segmentation with transformer},
  author={Wu, Junde and Ji, Wei and Fu, Huazhu and Xu, Min and Jin, Yueming and Xu, Yanwu},
  booktitle={Proceedings of the AAAI Conference on Artificial Intelligence},
  volume={38},
  number={6},
  pages={6030--6038},
  year={2024}
}

@inproceedings{wang2019boundary,
  title={Boundary and entropy-driven adversarial learning for fundus image segmentation},
  author={Wang, Shujun and Yu, Lequan and Li, Kang and Yang, Xin and Fu, Chi-Wing and Heng, Pheng-Ann},
  booktitle={Medical Image Computing and Computer Assisted Intervention--MICCAI 2019: 22nd International Conference, Shenzhen, China, October 13--17, 2019, Proceedings, Part I 22},
  pages={102--110},
  year={2019},
  organization={Springer}
}

@inproceedings{wenxuan2021transbts,
  title={Transbts: Multimodal brain tumor segmentation using transformer},
  author={Wenxuan, Wang and Chen, Chen and Meng, Ding and Hong, Yu and Sen, Zha and Jiangyun, Li},
  booktitle={International Conference on Medical Image Computing and Computer-Assisted Intervention, Springer},
  pages={109--119},
  year={2021}
}

@article{jiang2022swinbts,
  title={SwinBTS: A method for 3D multimodal brain tumor segmentation using swin transformer},
  author={Jiang, Yun and Zhang, Yuan and Lin, Xin and Dong, Jinkun and Cheng, Tongtong and Liang, Jing},
  journal={Brain sciences},
  volume={12},
  number={6},
  pages={797},
  year={2022},
  publisher={MDPI}
}

@article{isensee2021nnu,
  title={nnU-Net: a self-configuring method for deep learning-based biomedical image segmentation},
  author={Isensee, Fabian and Jaeger, Paul F and Kohl, Simon AA and Petersen, Jens and Maier-Hein, Klaus H},
  journal={Nature methods},
  volume={18},
  number={2},
  pages={203--211},
  year={2021},
  publisher={Nature Publishing Group}
}

@inproceedings{hatamizadeh2021swin,
  title={Swin unetr: Swin transformers for semantic segmentation of brain tumors in mri images},
  author={Hatamizadeh, Ali and Nath, Vishwesh and Tang, Yucheng and Yang, Dong and Roth, Holger R and Xu, Daguang},
  booktitle={International MICCAI brainlesion workshop},
  pages={272--284},
  year={2021},
  organization={Springer}
}

@inproceedings{wang2021boundary,
  title={Boundary-aware transformers for skin lesion segmentation},
  author={Wang, Jiacheng and Wei, Lan and Wang, Liansheng and Zhou, Qichao and Zhu, Lei and Qin, Jing},
  booktitle={Medical Image Computing and Computer Assisted Intervention--MICCAI 2021: 24th International Conference, Strasbourg, France, September 27--October 1, 2021, Proceedings, Part I 24},
  pages={206--216},
  year={2021},
  organization={Springer}
}

@article{wu2022fat,
  title={FAT-Net: Feature adaptive transformers for automated skin lesion segmentation},
  author={Wu, Huisi and Chen, Shihuai and Chen, Guilian and Wang, Wei and Lei, Baiying and Wen, Zhenkun},
  journal={Medical image analysis},
  volume={76},
  pages={102327},
  year={2022},
  publisher={Elsevier}
}

@article{maji2022attention,
  title={Attention Res-UNet with Guided Decoder for semantic segmentation of brain tumors},
  author={Maji, Dhiraj and Sigedar, Prarthana and Singh, Munendra},
  journal={Biomedical Signal Processing and Control},
  volume={71},
  pages={103077},
  year={2022},
  publisher={Elsevier}
}

@article{yang2017dagan,
  title={DAGAN: deep de-aliasing generative adversarial networks for fast compressed sensing MRI reconstruction},
  author={Yang, Guang and Yu, Simiao and Dong, Hao and Slabaugh, Greg and Dragotti, Pier Luigi and Ye, Xujiong and Liu, Fangde and Arridge, Simon and Keegan, Jennifer and Guo, Yike and others},
  journal={IEEE transactions on medical imaging},
  volume={37},
  number={6},
  pages={1310--1321},
  year={2017},
  publisher={IEEE}
}

\end{document}